\documentclass[conference,letter]{IEEEtran}
\usepackage{enumitem}
\usepackage{cite}
\usepackage{amsmath,amssymb,amsfonts}
\usepackage[ruled,vlined]{algorithm2e}
\usepackage{graphicx}
\usepackage{textcomp}
\usepackage{xcolor}
\usepackage{multirow}
\usepackage{booktabs}
\usepackage{subcaption}
\usepackage{hyperref}
\usepackage{todonotes}
\usepackage{color,soul}
\usepackage{xcolor}

\hypersetup{
     colorlinks = true,
     linkcolor = blue,
     anchorcolor = blue,
     citecolor = green,
     filecolor = magenta,
     urlcolor = cyan
     }

\ifCLASSINFOpdf
\else
\fi
\begin{document}
%

\title{\vspace{+1.0cm}Channel Boosting Feature Ensemble for Radar-based Object Detection}
\author{\IEEEauthorblockN{Shoaib Azam, Farzeen Munir and Moongu Jeon}
\IEEEauthorblockA{School of Electrical Engineering and Computer Science\\
Gwangju Institute of Science and Technology
Gwangju, South Korea\\
Email: {(shoaibazam, farzeen.munir, mgjeon)@gist.ac.kr}
}
}



%


\maketitle

\begin{abstract}
Autonomous vehicles are conceived to provide safe and secure services by validating the safety standards as indicated by SOTIF-ISO/PAS-21448 (Safety of the intended functionality)\footnote{https://www.daimler.com/innovation/case/autonomous/safety-first-for-automated-driving-2.htm}. Keeping in this context, the perception of the environment plays an instrumental role in conjunction with localization, planning and control modules. As a pivotal algorithm in the perception stack, object detection provides extensive insights into the autonomous vehicle's surroundings. Camera and Lidar are extensively utilized for object detection among different sensor modalities, but these exteroceptive sensors have limitations in resolution and adverse weather conditions. In this work, radar-based object detection is explored provides a counterpart sensor modality to be deployed and used in adverse weather conditions. The radar gives complex data; for this purpose, a channel boosting feature ensemble method with transformer encoder-decoder network is proposed. The object detection task using radar is formulated as a set prediction problem and evaluated on the publicly available dataset\cite{a9} in both good and good-bad weather conditions. The proposed method's efficacy is extensively evaluated using the COCO evaluation metric, and the best-proposed model surpasses its state-of-the-art counterpart method by $12.55\%$ and $12.48\%$ in both good and good-bad weather conditions.
\end{abstract}


%
\IEEEpeerreviewmaketitle

\section{Introduction}


The research and technological advancements made in the past three decades have enabled autonomous vehicles to become a certainty \cite{z4}. The development of sensors and computing technology has resulted in small size and cost-effective hardware for the autonomous vehicle. The sensor's accuracy, resolution, and latency have improved over the years to be consolidated in autonomous vehicles. The dynamic perception of the environment is fundamental to the safety of the autonomous vehicle. A broad set of exteroceptive sensors is utilized in the autonomous vehicle's hardware suite to perceive the environment. The typical sensor modalities include the camera, Lidar, radar, and sonar help the autonomous vehicle attempt to map, understand, and navigate the environment \cite{z2}. Among other perception tasks, the autonomous vehicle's most fundamental task is to identify and understand the objects encompassing it. 
\par


The most common exteroceptive sensor modality that is widely deployed for the perception of autonomous vehicles is cameras. Cameras operate in the visible spectrum and provide high-resolution information about the environment, which are essential to identify traffic lights, lane marking, and road obstacles. Besides providing in-depth details of the surroundings, cameras are sensitive to adverse weather conditions, illumination, and sun-glare, resulting in low-level image data that impediment the perception's accuracy for the autonomous vehicles. On the other hand, Lidar provides the $3$D information about the environment as a surrogate to $2$D details from the cameras. Lidar uses invisible laser light to measure the accurate distance of the objects. It measures nearly thousands of points to develop a $3$D point cloud map of the senor's surroundings. Despite providing the $3$D information, Lidar has a detriment of being expensive and also sensitive to adverse weather conditions. Contrary to cameras and Lidar, radar is less prone to adverse weather conditions and is used as a sensor modality for detecting small hazard objects in the autonomous vehicle sensor suite.  Most of the radar operates on the principle of Doppler's effect by firing radio waves at the target area and analyzing the reflected wave's frequency to estimate the object's speed and position. These type of radars mostly find their application in object tracking for the autonomous vehicle. Besides, there are scanning radars that provide the high resolution $360\deg$ range-azimuth images and are used for object detection in the perception stack of autonomous vehicles. Moreover, they are relatively cost-effective as compared to Lidar \cite{z1}. 
\par

Object detection forms the basis for the perception module and is formally done by using cameras and Lidar. A lot of research is carried out on the camera-based object detection that includes many state-of-the-art object detectors, for instance, YOLO-v3 \cite{yolov3}, SSD \cite{ssd}, Faster-RCNN \cite{faster}, RefineDet \cite{refine}, M2Det \cite{m2det}. These object detectors have limitation to operate in the adverse weather conditions \cite{z3}. In addition, much research is also focused on the use of Lidar for object detection. For this purpose, the Lidar data is represented either in rasterized \cite{zhou2018voxelnet} or geometric format \cite{qi2017pointnet} \cite{azam2018object} but have limitation to processing computation and adverse weather condition.  Besides, camera and Lidar object detection, there is still a room of improvement for the radar-based object detection. In this work, radar-based object detection in different weather condition is explored without utilizing any auxiliary sensor modality information. The radar-based object detection is formulated as a set prediction problem by designing a proposed network that includes channel boosting feature ensemble followed by transformer encoder-decoder architecture. The radar's data is complex; to determine rich feature representation, the channel boosting scheme is adopted in which the input radar image is transformed into different color space. The input radar image and color space images are followed by the respective backbone network for the features generation. The respective backbone networks' features are concatenated, and these image features are fed to transformer encoder-decoder architecture for object detection. The proposed network is extensively evaluated using the COCO evaluation metrics. The experimental evaluation shows the efficacy of proposed work with the state-of-the-art method \cite{a9}.  The main contributions are listed as follows:

\begin{figure*}[t]
      \centering
      \includegraphics[width=18cm]{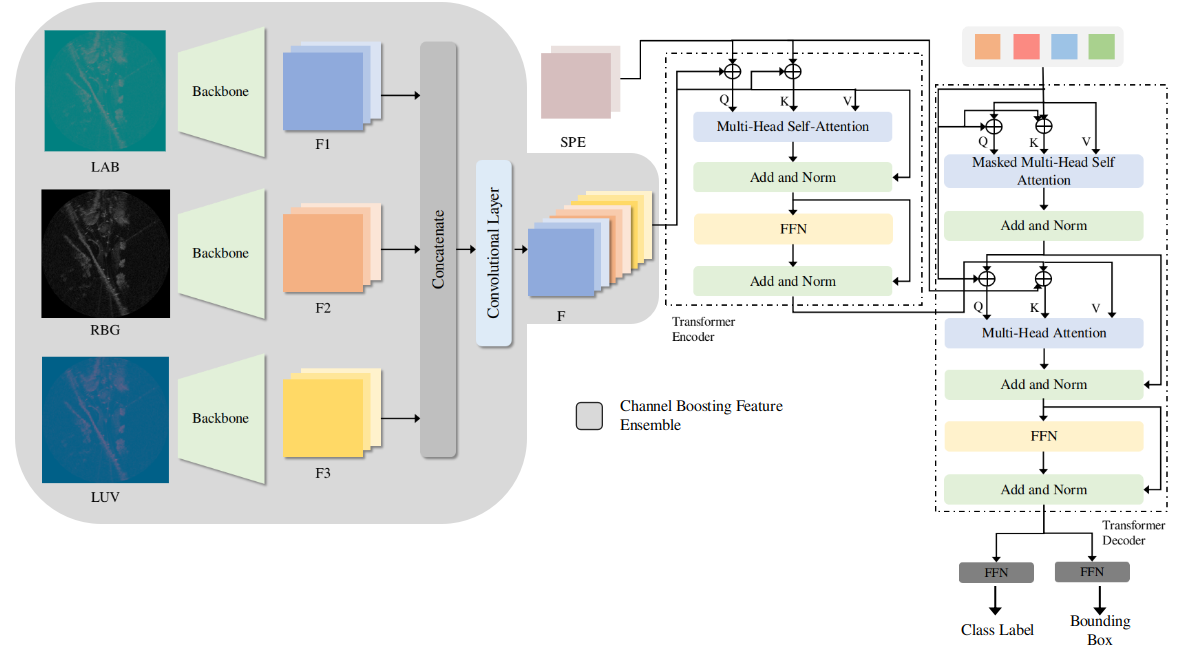}
      \caption{The overall framework for the proposed method. It includes a channel boosting feature ensemble method along with transformer encoder-decoder network for the radar-based object detection.}
      \label{framework}
\end{figure*}
\begin{enumerate}
    \item A novel channel boosting feature ensemble method is designed for the feature representation of radar data.
    \item The channel boosting feature ensemble method is used in conjunction with the transformer encoder-decoder architecture for the radar-based object detection.
    \item The efficacy of the proposed method (best-model) illustrates $12.55\%$ and $12.48\%$ increase in mean average precision (mAP) score in both good and good-bad weather conditions for radar-based object detection as compared to state-of-the-art (best-model) \cite{a9}.
\end{enumerate}

The rest of the paper is organized as follows: Section II gives related work. Section III explains the proposed methodology. The experimentation and results are discussed in Section IV, and section V concludes the paper.

\section{Related Work}
The researchers in \cite{k1,k2,k3} have explored radar-based object detection in the domain of autonomous vehicle. \cite{a1} has proposed a post-processing algorithm for object detected by highly accurate and reliable radar. The algorithm clusters multiple detections form a single object and track them using Kalman filter. \cite{a2} handles the uncertainties of laser radar by occupancy grid framework. The occupancy grid map is formed by using the inverse sensor model in the cartesian coordinate framework. The local grid map and global grid map, generated by temporal integration of sensor data are fused using the Dempster rule of combination. The conflict information is used to detect the moving object. 
\par
The fusion of radar and camera data for object detection is  done in \cite{g1,g2,g3}.  \cite{a3} has proposed the architecture for frontal object detection using radar data and a single camera. The detections are fused and then tracked using the Kalman filter. \cite{a4} represents a framework to identify an occupying area by fusing the data from millimetre radar and video camera. \cite{a5} performs the radar vision fusion in three steps. First radar guides a selection of the small number of candidates images for processing. Second, sparse representation is generated by normalizing the selected attention window and is processed by orientation-selective feature detector. Finally, a learned multi-layered network is used to classify the sparse representation of different objects. \cite{a6} has proposed a multi-sensor fusion system for object detection and tracking for autonomous vehicles. They have utilized the camera, lidar and radar to perceive the environment and develop a system to detect pedestrian, bicyclist and moving objects. The recognition framework helps to improve tracking and data association of the detected objects by multi-sensor data. 
\par
The aforementioned algorithm uses traditional techniques for object detection. Since the advent of the convolutional neural network (CNN), object detection accuracy has dramatically increased. CNN learns informative feature representation of objects contributes to success. In literature, convolutional neural networks are used for radar-based object detection. \cite{a7} has proposed a two-step object detection framework called radar region proposal network. The region proposal network generates a proposal by mapping radar data onto images and generates a set number of anchor boxes used to detect objects and calculate their distance from the vehicle. \cite{a8}  introduces an architecture for object detection that utilizes RGB camera images and radar sensor data. They have used radar data to identify high confidence point in the image to develop better feature representation and generate proposals for object detection. \cite{a9} has proposed radar dataset in adverse weather conditions and detect vehicles using radar data. They have used Faster-RCNN architecture for object detection with two modification. First, they fixed the anchor size and second, the bounding box has an additional parameter angle.  
\par
A new paradigm of the object detection algorithm has recently evolved based on a new attention-based building block called transformer network. The attention mechanism in transformers aggregates information of the whole input sequence, which give them the advantage of global computations and perfect memory. \cite{a10} introduces an encoder-decoder transformer framework for object detection based on a direct set prediction which bypasses anchor generation in typical object detection network.

\section{Proposed Method}
Feature extraction plays an imminent role for many deep-learning-based object detection task. In this work, a channel boosting feature ensemble method along with a direct set prediction for object detection using an encoder-decoder transformer network is proposed for the radar-based object detection. Fig.\ref{framework} illustrates the architecture of the proposed work.

\subsection{Channel Boosting Feature Ensemble Method}
The radar data exhibits large variations and thus requires a robust feature extraction mechanism. In this study, a channel boosting in conjunction with the ensemble method is used to extract features for radar images. First, the channel boosting method is used by aggregating the input image's RGB channels by converting it to LUV and CIELAB color spaces. The conversion to these color spaces enables the channel boosting by providing a good representation of the input image and helps the network to learn the modular and hierarchical representation of the image. Fig.\ref{framework} illustrates the channel boosting ensemble method for the representation of the input image. Suppose $x_i \in \mathbb{R}^{D \times H \times W}$ for $i \in I=(RGB,LUV,CIELAB)$ is the input image with $x_{i_{rgb}}$, $x_{i_{luv}}$ and $x_{i_{lab}}$ corresponds to the variants of the input image. Each channel variant of input image (RGB, LUV, and CIELAB) is followed by the convolution neural network (CNN) backbone for feature representation. Suppose the $x_{i\in rgb}$ with $D = 3$ (initially), is passed to the backbone for the generation of low-resolution activation map $f = \mathbb{R}^{C \times H \times W}$ with typical values of $C = 2048$. Similarly, the other two color space variants of the input image are processed by the backbone network resulting in low-resolution feature maps of each dimension of $C=2048$ respectively. These low-resolution feature activation maps are concatenated and then followed by an additional convolutional layer to produce the feature activation maps from the channel boosting ensemble method with a dimension of $C =2048$. The reason for this additional convolutional layer is empirical and also to reduce the computation cost.

\subsection{Transformer Encoder-Decoder}

\subsubsection{Encoder}
The transformer encoder expects a sequence as inputs, so the set of activation maps $f$ produced by channel boosting feature ensemble is reduced to lower dimension space from $C$ to $k$ by applying a $1 \times 1$ convolution layer and flattened to create the new feature map $\tilde{f} \in \mathbb{R}^{k \times HW}$ and is given to the transformer encoder. For each encoder layer, a multi-head self-attention module and the feed-forward network is incorporated in the architecture. 
\par 
For the multi-head self-attention module's brevity, the explanation of the single head attention mechanism is essential \cite{a10}\cite{vaswani2017attention}. The query and key-value pair form the basis of the attention and are utilized to map it to the output. The query, key and value are all represented in the vector format. The weight tensor $W_t \in \mathbb{R}^{3 \times \hat{k} \times k}$ for a single attention head $A (X_q, X{k_v},W_t)$ computes the query, key and value embedding after adding the query and key positional encoding ($P_q \in \mathbb{R}^{k \times N_q} $ and $P_{kv} \in \mathbb{R}^{k \times N_{kv}}$) respectively, as shown in Eq.\ref{equ1}
\begin{flalign}
\label{equ1}
\hspace{-1cm}
    \begin{aligned}
    &[Q:K:V] = [W_{t1}(X_q+P_q):W_{t_2}(X_{kv}+P_{kv}) \\
    &:W_{t_3}X_{kv}]
    \end{aligned}
\end{flalign}
where, $[:]$ represents the concatenation, and ($W_{t_1}$,$W_{t_2}$ and $W_{t_3}$) are concatenation of $W_t$. $X_q$ corresponds to the query sequence of length $N_q$. The key-value sequence is denoted by $X_{k_v}$ of length $N_{k_v}$.  Eq.\ref{equ2} illustrates the computation of activation weights $\delta$ by computing the soft-max function of the dot product of queries and keys.
\begin{align}
 \label{equ2}
 \hspace{-4cm}
   \delta_{(i,j)} = \frac{\exp^{\frac{1}{\sqrt{k}}Q_{i}^{T}K_j}}{\sum_{j=1}^{N_{kv}}\exp^{\frac{1}{\sqrt{k}}Q_{i}^{T}K_j}}
\end{align}
The positional embedding is learned and shared across all the attention layers for the given query, key-value pair \cite{a10}. The final output of the single head attention module is computed by Eq.\ref{equ3}.
\begin{flalign}
    \label{equ3}
     \hspace{-3.5cm}
    A(X_q,X_{kv},W) = \sum_{j=1}^{N_{kv}}\delta_{ij}V_j
\end{flalign}
The multi-head self-attention module is the concatenation of $M$ single head attention modules followed by the projection matrix $L$. The projection matrix is the combination of residual connection, dropout and layer normalization. The overall computation is described in Eq.\ref{equ4}. In the case of multi-head attention module the only difference in computing the attention weight as described in Eq.\ref{equ2} is the change of scale fraction $k$ to $\hat{k}=\frac{k}{M}$. 
\begin{flalign}
    \label{equ4}
    \hspace{-0.5cm}
    \begin{aligned}
        &{X_q}' = [A(X_q,X_{kv},W_1): ... :A(X_q,X_{kv},W_M)] \\
    &\tilde{X_q}= \psi (X_q+dropout(L{X_q}')) 
    \end{aligned}
\end{flalign}

\subsubsection{Transformer Decoder}
The decoder follows the same structure of sub-layer as the encoder. It constitutes of two multi-head self-attention modules used for transforming the $N$ embedding size of $k$. 
\par
The decoder's input are queries that are initially set to zero, $N$ object queries that are learnt positional encodings, and the encoder memory. In the decoder, the positional encodings are added to each attention layers. The decoder's aforementioned inputs are used to produce the final set of prediction that includes bounding box and class labels through multi-head self-attention decoder modules. It is to be mentioned here that the first self-attention layer in the first decoder layer is skipped in the processing of decoding the embeddings.

\subsection{Feed-Forward Network (FFN)}
The decoder's output embedding is fed to the feed-forward neural network for the prediction of class labels and bounding box.  A $3$ layer perceptron network with ReLU activation, a hidden dimension of $k$ and a linear projection layer is used for the final prediction output. The output of FFN consists of height, the width of the bounding box w.r.t image, normalized center coordinates and class labels. Besides, an extra no-class label is also used for no object detected in the input image. In this work, the number of classes is $2$ (vehicle and no-vehicle) because the intra-class object detection will not provide any sufficient information because of the radar data nature.

\section{Experimentation and Results}

\subsection{Dataset}
RADIATE (RAdar Dataset In Adverse weaThEr) dataset is used in this study \cite{a9}. The motivation to use RADIATE is to facilitate the object detection research in adverse weather conditions and understand the dynamic environment better so that an autonomous vehicle can safely navigate. The dataset uses Navtech CTS350-X radar to collect the data.  The radar renders  $360^{\circ}$ high-resolution range-azimuth images. It has a maximum operating range of 100 meters and 0.175m range resolution, $1.8^{\circ}$ azimuths and elevation resolution. 
The dataset is collected in different weather conditions including sunny, overcast, night, snow, fog and rain. The dataset consists of $3$ hours of annotated radar images with an estimated $200$K labels vehicles. The driving scenarios covers urban, motorway and suburban driving. The images are labelled using 2D bounding boxes. Each bounding box define $(x,y, width, height, angle)$, where $(x,y)$ are the top-left pixel location and angle define counter-clockwise rotation.   The dataset is divided into three parts, training in good weather, training in good-bad weather and test set that contain both good and bad weather. We have adopted standard partition as proposed by \cite{a9} for a fair comparison, shown in Table \ref{table3}.
\begin{table}[]
\centering
\caption{The RADIATE dataset partition topology for training, and testing the proposed method in adverse weather conditions }
\label{table3}
\begin{tabular}{@{}l|c|c@{}}
\toprule
 & \multicolumn{1}{l|}{No. of Images} & \multicolumn{1}{l}{No. of vehicles} \\ \midrule
Train Set (Good Weather)         & 23091 & 106931 \\
Train Set (Good and Bad Weather) & 9760  & 39647  \\
Test Set                             & 11289 & 147005 \\ \bottomrule
\end{tabular}
\end{table}

\subsection{Evaluation Metric}
Numerous evaluation metrics exist to evaluate the accuracy of object detection in the images. Here, we have used a standard MS COCO evaluation metric \cite{coco}. IoU (Intersection over Union) defined the intersection between the area of ground-truth bounding box and area of the predicted bounding box shown by Eq. \ref{equa2}. The prediction is True Positive (TP) if $IoU > threshold$ and False Positive (FP) if $IoU < threshold$. 
\begin{equation}
    \begin{aligned}
    \label{equa2}
    \hspace{-3.7cm}
   IoU= \frac{A_p \cap A_{gt}}{A_p \cup A_{gt}}
\end{aligned}
\end{equation}

Recall and Precision are calculated using Eq. \ref{equa3}.
\begin{equation}
   \begin{aligned}
    \label{equa3}
     \hspace{-0.3cm}
   & Recall =\frac{TP}{TP+FN}, Precision =\frac{TP}{TP+FP}
\end{aligned} 
\end{equation}

The Average Precision(AP) is calculated per class. AP is the area under the PR curve, shown by Eq. \ref{equa1}.
\begin{equation}
   \begin{aligned}
    \label{equa1}
    \hspace{-0.2cm}
   AP[class]=\frac{1}{\#thresh} \sum_{IoU \in thresh} AP[class,IoU]
\end{aligned} 
\end{equation}

COCO evaluation uses a threshold range from $0.5$ to $0.95$ with a step size of $0.05$. In the experimentation, $mAP_{IoU=0.5}$ is utilized, which is also similar to PASCAL VOC \cite{everingham2010pascal} evaluation.

\begin{figure}[t]
      \centering
      \includegraphics[width=8cm]{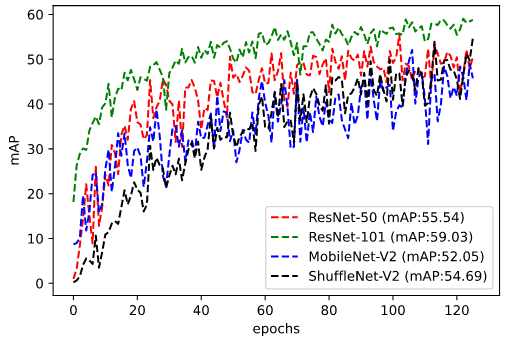}
      \caption{The illustration of mAP scores for all the models used as a backbone network in the proposed method trained on good-bad weather conditions and tested on test data. ResNet-$101$ has the best results. }
      \label{good-bad}
\end{figure}

\begin{figure}[t]
      \centering
      \includegraphics[width=8cm]{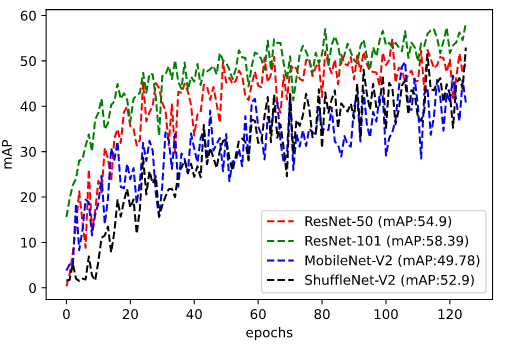}
      \caption{The illustration of mAP scores for all the models used as a backbone network in the proposed method trained on good weather conditions and tested on test data. ResNet-$101$ has the best results.}
      \label{good}
\end{figure}

\subsection{Experimentation}
The proposed network is trained on two GPUs having $24Gb$ memory in total using Pytorch deep learning library. The input training data is scaled, augmented, randomly horizontal flipped, randomly resized and randomly cropped in order to provide the network to learn the global relationship. The training process is run for $125$ epochs having the learning rate drop of factor $10$ at $100$ epochs. The Adam (Adaptive Momentum Estimation) is used as an optimizer with $10^{-5}$ and $10^{-4}$ learning rates for the backbone and transformer network, respectively. The pre-trained backbone network is imported from the Torchvision with frozen the batch-norm layer and discarding the last classification layer. In the experimentation, the ResNet-$50$ \cite{he2015deep}, RestNet-$101$ \cite{he2015deep}, MobileNet-V$2$ \cite{sandler2018mobilenetv2} and ShuffleNet-V$2$ \cite{ma2018shufflenet} are used as a backbone networks. In the transformer network, the number of encoders and decoders are equal to $6$. All the transformer weights are randomly initialized with Xavier initialization. A dropout of $0.1$ is used after every multi-head attention and feed-forward network layer before normalization. For the spatial positional encodings, an absolute position encodings are used that are function of sine and cosine with different frequencies and are concatenated to get the final position encoding across the $k$ channel. 

\begin{figure*}[t]
      \centering
      \includegraphics[width=13cm]{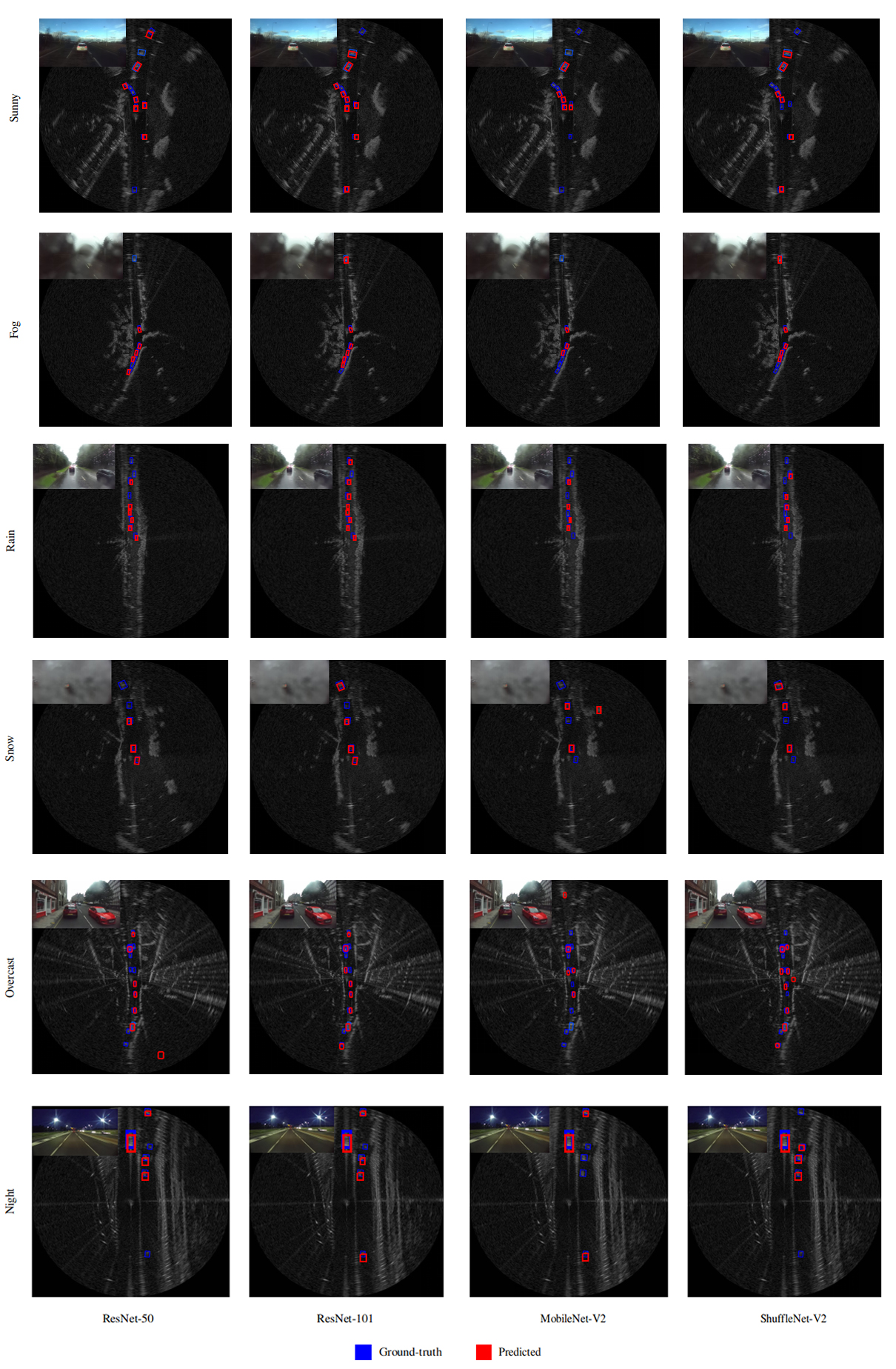}
      \caption{The qualitative results of proposed radar-based object detection in different weather conditions with all the four backbone networks used along with transformer encoder-decoder network is illustrated. }
      \label{results}
\end{figure*}

\par 
Since, the proposed method infers the set of $N$ predictions, in this work, inspired by \cite{a10}  an optimal bipartite matching scheme is adopted for the loss calculations. Eq.\ref{equ6} illustrates the bipartite matching loss function between $\hat{y}$ set of predictions $N$ added with no object class and ground-truth ($y$) also padded with the set of size $N$.

\begin{align}
    \label{equ6}
    \hspace{-2.5cm}
    \tilde{\eta} = \underset{\eta \in C_N}{argmin} \sum_{j}^{N}L_{match}(y_j,\hat{y}_{\eta(j)})
\end{align}
where $C_N$ corresponds to the permutations of N elements. $L_{match}(y_j,\hat{y}_{\eta(j)})$ illustrates the pair-wise matching cost between ground-truth and prediction. A Hungarian algorithm is used to compute this assignment problem between ground-truth and prediction. The matching cost constitutes towards both the class labels and the bounding box between the ground-truth and the predictions. Suppose each element $j$ of ground-truth set is represented as $y_j = (c_j,b_j)$ where $c_j$ is the class label and $b_j \in [0,1]^5$ is the bounding box vector corresponds to normalized center coordinates, height, width and angle for the respective image. For the prediction, the class probability and bounding box is defined as $\hat{p}_{\eta{j}}(c_j)$ and $\hat{b}_{\eta{j}}$ respectively. The matching between ground-truth and prediction is done by finding the direct one-to-one correspondence without duplicates. The modified Hungarian loss function with angle inclusion in the bounding box is illustrated in Eq.\ref{equ7} using the aforementioned notations of the class label and bounding box.
\begin{flalign}
    \label{equ7}
    \hspace{-2cm}
    \begin{aligned}
    L_{Hungarian}(y,\hat{y})= \sum_{j=1}^{N}[-log\hat{p}_{\eta({j})}(c_j)+ & \\ \mathbf{1}_{c_j\neq(\phi)}L_{box}(b_j,\hat{b}_{\hat{\eta}}(j))] 
    \end{aligned}
\end{flalign}
where $\hat{\eta}$ corresponds to optimal assignment. The $L_{box}(.)$ in the Hungarian loss function is used for scoring the bounding box, and in comparison to other detectors the box prediction is computed directly. In other to compensate the scaling issue, the $l_1$ loss with the complete IoU loss is used. The complete IoU \cite{zheng2020distance} loss is illustrated by the Eq.\ref{equ8}. 
    

\begin{align}
    \label{equ8}
    \hspace{-2.4cm}
   L_{CIoU}=1-IoU+\frac{\rho (\mathbf{b},\boldsymbol{b_{gt}})}{c^2}+\alpha \nu 
\end{align}
where  $1-IoU+\frac{\rho (\mathbf{b},\boldsymbol{b_{gt}})}{c^2}$ corresponds to distance IoU with $\mathbf{b}$, $\boldsymbol{b_{gt}}$ denotes the bounding box points of predicted and ground-truth respectively. $\rho$ denotes the euclidean distance and $c$ illustrates the diagonal length of smallest enclosing box that covers the two boxes. $\alpha$ is the positive trade-off parameter and the $\nu$ is used to compute the consistency  of aspect ratio defined by Eq.\ref{equ9} and Eq.\ref{equ10} respectively.

\begin{equation}
 \begin{aligned}
    \label{equ9}
    \hspace{-3.9cm}
   \alpha = \frac{\nu}{(1-IoU)+\nu}
\end{aligned}   
\end{equation}

\begin{equation}
    \begin{aligned}
    \label{equ10}
    \hspace{-2.8cm}
   \nu = \frac{4}{\pi^2}(arctan\frac{w^{gt}}{h^{gt}}-arctan\frac{w}{h})^2
\end{aligned}
\end{equation}

The $L_{box}(.)$ is represented as shown in Eq.\ref{equ11}
\begin{equation}
\label{equ11}
\hspace{-1.9cm}
    \begin{aligned}
    L_{box}(b_j,\hat{b}_{\hat{\eta}}(j)) = \lambda_{iou}L_{CIoU}(b_j,\hat{b}_{\hat{\eta}}(j)) &+ \\
    \lambda_{l1}\left \|b_j- \hat{b}_{\hat{\eta}}(j) \right \|
    \end{aligned}
\end{equation}

where $\lambda_{l1}=4$ and $\lambda_{iou}=2$ are used in training the proposed method.
 \clearpage

\par
For the quantitative and qualitative evaluation, a baseline network of proposed work is trained on two sets of datasets that include good-bad weather conditions and only good weather conditions. The proposed baseline network is compared with \cite{a9} for the evaluation to measure the mean average precision (mAP) scores. Table-\ref{table1} and Table-\ref{table2} illustrates the mAP score of proposed baseline method on testing data. In addition, to further explore the proposed method's effectiveness, the proposed method is trained using RestNet-$101$, MobileNet-V$2$ and ShuffleNet-V$2$ as a backbone network. In the quantitative analysis, ResNet-$101$ has outperformed the other backbone network mAP scores and, also surpasses the \cite{a9} scores by a margin of $12.48\%$ in good-bad weather conditions and by $12.55 \%$ in good weather conditions with its counterpart network. Fig.\ref{good-bad} and Fig.\ref{good} illustrates the running mAP scores of good-bad weather conditions and good weather conditions across the number of epochs on the testing data. Fig.\ref{results} shows the qualitative results of radar object detection of all the backbone used in the experimentation in different weather conditions along with the accompanying optical image of the scene.

\begin{table}[]
\centering
\caption{mAP Score on Radar Data with Good and Bad Weather Conditions}
\label{table1}
\begin{tabular}{@{}l|c@{}}
\toprule
Models                                       & \multicolumn{1}{l}{mAP Score} \\ \midrule
ResNet-50 Trained \cite{a9}                  & 45.77                         \\
ResNet-101 \cite{a9}                         & 46.55                         \\
Ours (Backbone: ResNet-50) (Baseline)         & 55.54                         \\
Ours (Backbone: ShuffleNet-V2)                & 54.69                         \\
Ours (Backbone: MobileNet-V2)                 & 52.05                         \\
\textbf{Ours (Backbone: ResNet-101)}          & \textbf{59.03}                \\ \bottomrule
\end{tabular}
\end{table}

\begin{table}[]
\centering
\caption{mAP Score on Radar Data with Good Weather Conditions}
\label{table2}
\begin{tabular}{@{}lc@{}}
\toprule
\multicolumn{1}{l|}{Models}                                     & mAP Score            \\ \midrule
\multicolumn{1}{l|}{ResNet-50 Trained \cite{a9}}                & 45.31                \\
\multicolumn{1}{l|}{ResNet-101 \cite{a9}}                        & 45.84                \\
\multicolumn{1}{l|}{Ours (Backbone: ResNet-50) (Baseline)}       & 54.90                 \\
\multicolumn{1}{l|}{Ours (Backbone: ShuffleNet-V2)}              & 52.90                 \\
\multicolumn{1}{l|}{Ours (Backbone: MobileNet-V2)}               & 49.78                \\
\multicolumn{1}{l|}{\textbf{Ours (Backbone: ResNet-101)}}                 & \textbf{58.39}                \\ \midrule
                                                                & \multicolumn{1}{l}{}
\end{tabular}
\end{table}

\section{Conclusion}
This study focuses on radar-based object detection in adverse weather condition for autonomous vehicles. A novel architecture has introduced which utilizes the channel boosting feature ensemble for more desirable feature extraction. The extracted features are then concatenated and feed to the encoder-decoder transformer network, which uses a direct set prediction method for object detection. The efficacy of the proposed algorithm is evaluated using standard COCO evaluation. The mean average precision of $59.03\%$ and $58.39\%$ are achieved on test data for the proposed method with ResNet-$101$ backbone netowrk trained on good-bad weather conditions and good weather conditions respectively. 
\par
In future work, we aim to fuse Lidar data with radar to improve object detection in adverse weather condition. Moreover, use image data to classify the weather condition to optimize the perception system of the autonomous vehicle.


\section*{Acknowledgment}
This work was partly supported by Institute of Information communications Technology Planning Evaluation (IITP) grant funded by the Korea Government (MSIT) (No. 2014-3-00077, Development of Global Multi-target Tracking and Event Prediction Techniques Based on Real-time Large-Scale Video Analysis) and the National Research Foundation of Korea (NRF) grant funded by the Korea Government (MSIT) (No. 2019R1A2C2087489), and Ministry of Culture, Sports and Tourism (MCST), and Korea Creative Content Agency (KOCCA) in the Culture Technology (CT) Research \& Development (R2020070004) Program 2020.





\bibliographystyle{IEEEtran}
\bibliography{ref.bib}

\begin{thebibliography}{10}
\providecommand{\url}[1]{#1}
\csname url@samestyle\endcsname
\providecommand{\newblock}{\relax}
\providecommand{\bibinfo}[2]{#2}
\providecommand{\BIBentrySTDinterwordspacing}{\spaceskip=0pt\relax}
\providecommand{\BIBentryALTinterwordstretchfactor}{4}
\providecommand{\BIBentryALTinterwordspacing}{\spaceskip=\fontdimen2\font plus
\BIBentryALTinterwordstretchfactor\fontdimen3\font minus
  \fontdimen4\font\relax}
\providecommand{\BIBforeignlanguage}[2]{{%
\expandafter\ifx\csname l@#1\endcsname\relax
\typeout{** WARNING: IEEEtran.bst: No hyphenation pattern has been}%
\typeout{** loaded for the language `#1'. Using the pattern for}%
\typeout{** the default language instead.}%
\else
\language=\csname l@#1\endcsname
\fi
#2}}
\providecommand{\BIBdecl}{\relax}
\BIBdecl

\bibitem{a9}
M.~Sheeny, E.~De~Pellegrin, S.~Mukherjee, A.~Ahrabian, S.~Wang, and A.~Wallace,
  ``Radiate: A radar dataset for automotive perception,'' \emph{arXiv preprint
  arXiv:2010.09076}, 2020.

\bibitem{z4}
K.~Bimbraw, ``Autonomous cars: Past, present and future a review of the
  developments in the last century, the present scenario and the expected
  future of autonomous vehicle technology,'' in \emph{2015 12th international
  conference on informatics in control, automation and robotics (ICINCO)},
  vol.~1.\hskip 1em plus 0.5em minus 0.4em\relax IEEE, 2015, pp. 191--198.

\bibitem{z2}
S.~Campbell, N.~O'Mahony, L.~Krpalcova, D.~Riordan, J.~Walsh, A.~Murphy, and
  C.~Ryan, ``Sensor technology in autonomous vehicles: A review,'' in
  \emph{2018 29th Irish Signals and Systems Conference (ISSC)}.\hskip 1em plus
  0.5em minus 0.4em\relax IEEE, 2018, pp. 1--4.

\bibitem{z1}
F.~Rosique, P.~J. Navarro, C.~Fern{\'a}ndez, and A.~Padilla, ``A systematic
  review of perception system and simulators for autonomous vehicles
  research,'' \emph{Sensors}, vol.~19, no.~3, p. 648, 2019.

\bibitem{yolov3}
J.~Redmon and A.~Farhadi, ``Yolov3: An incremental improvement,'' \emph{arXiv
  preprint arXiv:1804.02767}, 2018.

\bibitem{ssd}
W.~Liu, D.~Anguelov, D.~Erhan, C.~Szegedy, S.~Reed, C.-Y. Fu, and A.~C. Berg,
  ``Ssd: Single shot multibox detector,'' in \emph{European conference on
  computer vision}.\hskip 1em plus 0.5em minus 0.4em\relax Springer, 2016, pp.
  21--37.

\bibitem{faster}
S.~Ren, K.~He, R.~B. Girshick, and J.~Sun, ``Faster r-cnn: towards real-time
  object detection with region proposal networks. corr abs/1506.01497 (2015),''
  \emph{arXiv preprint arXiv:1506.01497}, 2015.

\bibitem{refine}
S.~Zhang, L.~Wen, X.~Bian, Z.~Lei, and S.~Z. Li, ``Single-shot refinement
  neural network for object detection,'' in \emph{Proceedings of the IEEE
  conference on computer vision and pattern recognition}, 2018, pp. 4203--4212.

\bibitem{m2det}
Q.~Zhao, T.~Sheng, Y.~Wang, Z.~Tang, Y.~Chen, L.~Cai, and H.~Ling, ``M2det: A
  single-shot object detector based on multi-level feature pyramid network,''
  in \emph{Proceedings of the AAAI Conference on Artificial Intelligence},
  vol.~33, 2019, pp. 9259--9266.

\bibitem{z3}
S.~D. Pendleton, H.~Andersen, X.~Du, X.~Shen, M.~Meghjani, Y.~H. Eng, D.~Rus,
  and M.~H. Ang, ``Perception, planning, control, and coordination for
  autonomous vehicles,'' \emph{Machines}, vol.~5, no.~1, p.~6, 2017.

\bibitem{zhou2018voxelnet}
Y.~Zhou and O.~Tuzel, ``Voxelnet: End-to-end learning for point cloud based 3d
  object detection,'' in \emph{Proceedings of the IEEE Conference on Computer
  Vision and Pattern Recognition}, 2018, pp. 4490--4499.

\bibitem{qi2017pointnet}
C.~R. Qi, H.~Su, K.~Mo, and L.~J. Guibas, ``Pointnet: Deep learning on point
  sets for 3d classification and segmentation,'' in \emph{Proceedings of the
  IEEE conference on computer vision and pattern recognition}, 2017, pp.
  652--660.

\bibitem{azam2018object}
S.~Azam, F.~Munir, A.~Rafique, Y.~Ko, A.~M. Sheri, and M.~Jeon, ``Object
  modeling from 3d point cloud data for self-driving vehicles,'' in \emph{2018
  IEEE Intelligent Vehicles Symposium (IV)}.\hskip 1em plus 0.5em minus
  0.4em\relax IEEE, 2018, pp. 409--414.

\bibitem{k1}
C.~C. Schwesig and I.~Poupyrev, ``Radar-based object detection for vehicles,''
  Oct.~29 2019, uS Patent 10,459,080.

\bibitem{k2}
D.~M. Gavrila, ``Sensor-based pedestrian protection,'' \emph{IEEE Intelligent
  Systems}, vol.~16, no.~6, pp. 77--81, 2001.

\bibitem{k3}
S.~Miyahara, ``New algorithm for multiple object detection in fm-cw radar,''
  SAE Technical Paper, Tech. Rep., 2004.

\bibitem{a1}
A.~Manjunath, Y.~Liu, B.~Henriques, and A.~Engstle, ``Radar based object
  detection and tracking for autonomous driving,'' in \emph{2018 IEEE MTT-S
  International Conference on Microwaves for Intelligent Mobility
  (ICMIM)}.\hskip 1em plus 0.5em minus 0.4em\relax IEEE, 2018, pp. 1--4.

\bibitem{a2}
J.~Duan, L.~Ren, L.~Li, and D.~Liu, ``Moving objects detection in evidential
  occupancy grids using laser radar,'' in \emph{2016 8th International
  Conference on Intelligent Human-Machine Systems and Cybernetics (IHMSC)},
  vol.~2.\hskip 1em plus 0.5em minus 0.4em\relax IEEE, 2016, pp. 73--76.

\bibitem{g1}
M.~Gong, Z.~Zhou, and J.~Ma, ``Change detection in synthetic aperture radar
  images based on image fusion and fuzzy clustering,'' \emph{IEEE Transactions
  on Image Processing}, vol.~21, no.~4, pp. 2141--2151, 2011.

\bibitem{g2}
Y.-T. Zhou, ``Multi-sensor image fusion,'' in \emph{Proceedings of 1st
  International Conference on Image Processing}, vol.~1.\hskip 1em plus 0.5em
  minus 0.4em\relax IEEE, 1994, pp. 193--197.

\bibitem{g3}
K.~Abe, S.~Tokoro, and K.~Suzuki, ``Object detection system and method of
  detecting object,'' Apr.~15 2008, uS Patent 7,358,889.

\bibitem{a3}
R.~O. Chavez-Garcia, J.~Burlet, T.-D. Vu, and O.~Aycard, ``Frontal object
  perception using radar and mono-vision,'' in \emph{2012 IEEE Intelligent
  Vehicles Symposium}.\hskip 1em plus 0.5em minus 0.4em\relax IEEE, 2012, pp.
  159--164.

\bibitem{a4}
T.~Kato, Y.~Ninomiya, and I.~Masaki, ``An obstacle detection method by fusion
  of radar and motion stereo,'' \emph{IEEE transactions on intelligent
  transportation systems}, vol.~3, no.~3, pp. 182--188, 2002.

\bibitem{a5}
Z.~Ji and D.~Prokhorov, ``Radar-vision fusion for object classification,'' in
  \emph{2008 11th International Conference on Information Fusion}.\hskip 1em
  plus 0.5em minus 0.4em\relax IEEE, 2008, pp. 1--7.

\bibitem{a6}
H.~Cho, Y.-W. Seo, B.~V. Kumar, and R.~R. Rajkumar, ``A multi-sensor fusion
  system for moving object detection and tracking in urban driving
  environments,'' in \emph{2014 IEEE International Conference on Robotics and
  Automation (ICRA)}.\hskip 1em plus 0.5em minus 0.4em\relax IEEE, 2014, pp.
  1836--1843.

\bibitem{a7}
R.~Nabati and H.~Qi, ``Rrpn: Radar region proposal network for object detection
  in autonomous vehicles,'' in \emph{2019 IEEE International Conference on
  Image Processing (ICIP)}.\hskip 1em plus 0.5em minus 0.4em\relax IEEE, 2019,
  pp. 3093--3097.

\bibitem{a8}
R.~Yadav, A.~Vierling, and K.~Berns, ``Radar+ rgb attentive fusion for robust
  object detection in autonomous vehicles,'' \emph{arXiv preprint
  arXiv:2008.13642}, 2020.

\bibitem{a10}
N.~Carion, F.~Massa, G.~Synnaeve, N.~Usunier, A.~Kirillov, and S.~Zagoruyko,
  ``End-to-end object detection with transformers,'' \emph{arXiv preprint
  arXiv:2005.12872}, 2020.

\bibitem{vaswani2017attention}
A.~Vaswani, N.~Shazeer, N.~Parmar, J.~Uszkoreit, L.~Jones, A.~N. Gomez,
  {\L}.~Kaiser, and I.~Polosukhin, ``Attention is all you need,'' in
  \emph{Advances in neural information processing systems}, 2017, pp.
  5998--6008.

\bibitem{coco}
T.-Y. Lin, M.~Maire, S.~Belongie, J.~Hays, P.~Perona, D.~Ramanan,
  P.~Doll{\'a}r, and C.~L. Zitnick, ``Microsoft coco: Common objects in
  context,'' in \emph{European conference on computer vision}.\hskip 1em plus
  0.5em minus 0.4em\relax Springer, 2014, pp. 740--755.

\bibitem{everingham2010pascal}
M.~Everingham, L.~Van~Gool, C.~K. Williams, J.~Winn, and A.~Zisserman, ``The
  pascal visual object classes (voc) challenge,'' \emph{International journal
  of computer vision}, vol.~88, no.~2, pp. 303--338, 2010.

\bibitem{he2015deep}
K.~He, X.~Zhang, S.~Ren, and J.~Sun, ``Deep residual learning for image
  recognition. corr abs/1512.03385 (2015),'' 2015.

\bibitem{sandler2018mobilenetv2}
M.~Sandler, A.~Howard, M.~Zhu, A.~Zhmoginov, and L.-C. Chen, ``Mobilenetv2:
  Inverted residuals and linear bottlenecks,'' in \emph{Proceedings of the IEEE
  conference on computer vision and pattern recognition}, 2018, pp. 4510--4520.

\bibitem{ma2018shufflenet}
N.~Ma, X.~Zhang, H.-T. Zheng, and J.~Sun, ``Shufflenet v2: Practical guidelines
  for efficient cnn architecture design,'' in \emph{Proceedings of the European
  conference on computer vision (ECCV)}, 2018, pp. 116--131.

\bibitem{zheng2020distance}
Z.~Zheng, P.~Wang, W.~Liu, J.~Li, R.~Ye, and D.~Ren, ``Distance-iou loss:
  Faster and better learning for bounding box regression.'' in \emph{AAAI},
  2020, pp. 12\,993--13\,000.

\end{thebibliography}
%



\end{document}